\documentclass[letterpaper, 10 pt, conference]{ieeeconf}  

\IEEEoverridecommandlockouts                              
\usepackage{graphicx}
\usepackage{cite}
\usepackage{amsmath}
\usepackage{subcaption}

\title{\LARGE \bf
Identifying robust landmarks in feature-based maps
}

\author{Julie Stephany Berrio $^{1}$, James Ward $^{1}$, Stewart Worrall $^{1}$,   Eduardo Nebot$^{1}$
\thanks{$^{1}$J. Berrio, J. Ward, S. Worrall, E. Nebot  are with the Australian Centre for Field Robotics (ACFR) at the University of Sydney (NSW, Australia).
       E-mails: {\tt\small \{j.berrio, j.ward, s.worrall, e.nebot\}@acfr.usyd.edu.au}}
}

\begin{document}

\maketitle
\thispagestyle{empty}
\pagestyle{empty}

\begin{abstract}

To operate in an urban environment, an automated vehicle must be capable of accurately estimating its position within a global map reference frame.
This is necessary for optimal path planning and safe navigation.
To accomplish this over an extended period of time, the global map requires long term maintenance.
This includes the addition of newly observable features and the removal of transient features belonging to dynamic objects.
The latter is especially important for the long-term use of the map as matching against a map with features that no longer exist can result in incorrect data associations, and consequently erroneous localisation.
This paper addresses the problem of removing features from the map that correspond to objects that are no longer observable/present in the environment. 
This is achieved by assigning a single score which depends on the geometric distribution and characteristics when the features are re-detected (or not) on different occasions.
Our approach not only eliminates ephemeral features, but also can be used as a reduction algorithm for highly dense maps.
We tested our approach using half a year of weekly drives over the same 500 metre section of road in an urban environment. 
The results presented demonstrate the validity of the long term approach to map maintenance.

\end{abstract}

\section{Introduction}

The accurate estimation of vehicle pose within a global environment is a fundamental requirement for the navigation of an autonomous vehicle. Urban vehicles must be able to operate in areas where GNSS cannot provide accurate localisation, or at times does not provide any position information at all \cite{TFAD}. This means that an alternative method of localisation is required, and a popular approach is to perform pose estimation based on the mapping of distinctive landmarks in the operational area.

The generation of such maps is challenging as it requires the initial acquisition of sensor data, information analysis, and continuous maintenance \cite{kuutti}.
Feature-based maps provide a compact representation of the environment. The main assumption in their construction is that the world is static, and hence the features do not change over time. 
Also, when the map is first generated every feature observed by the perception system is incorporated as a map landmark, whether or not it represents a static or dynamic object. 
Consequently, maps should be updated as the environment changes to aid not only localisation, but motion planning in dynamic environments \cite{6907397}. 

The initial map features and their properties (dynamic/static, feature class) obtained from a SLAM type algorithm depend on the configuration parameters that control the sensitivity and quality of the feature detector.
For example, if the detector is restricted to selecting only features that are located above 3 meters from the ground plane, it is far more likely that the detected features correspond to static objects. 
The problem with using a very restrictive detector is that there is a corresponding reduction in the number of features. 
With fewer features to enable the localisation, there is an increased chance of poor map matching and failures in the data association procedure. 

The maintenance process generally includes long term data collection from a vehicle travelling within the same area on different days, times of day or seasons.
The post processing of information should identify and incorporate any environmental changes into the map representation. 
In a typical setup, a local map is created and stored for the current session while the vehicle is navigating \cite{5245983, 7358940, 8023504}. 
In the case that a known place is visited again, new observations are used to improve the map by adding new features \cite{M_dymczyk_2016} and incrementally updating the feature map.

\begin{figure}[t]
   \centering
   \includegraphics[scale=.33]{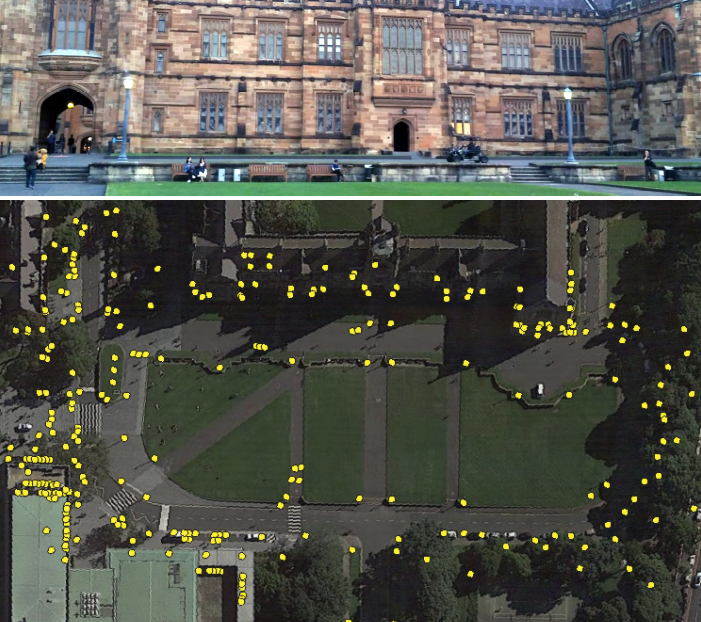}
    \caption{ Single-session feature map built with an electric vehicle platform equipped with a VLP-16 laser sensor to detect features in the environment (poles/corners).}
    \label{fig:mapa}
\end{figure}

For the operation of a SLAM type approach, incorporating more features into the state vector can lead to an unmanageable amount of information. 
This is compounded when using a map with features belonging to objects that are no longer in the scene. 
In some approaches, the map is stored and queried in order to retrieve only the features which could be associated with the current sensor information. 
When a system is bandwidth limited the amount of data to be transferred and computed has to be taken into account.
In both cases is important determine when the map features no longer correspond to objects in the environment, not only to keep the compactness and reliability of the map, but also to simplify the data association problem.

In this paper we present a method to update the information contained in a feature map using a process that estimates a feature score. This approach can also be used as a method to downsample the map.
A regression model is trained with a data set labelled using empirical probability estimation.
The data set consists of a group of identified predictors based on the information collected over 6 months using an electric vehicle. This vehicle is equipped with various sensors, and was driven to follow approximately the same trajectory once per week within a predefined area. 
A prior feature map was created from a dataset collected during a single drive around this area.

Our approach achieved both the scoring and selection of reliable features and also the downsampling of the map while maintaining a guaranteed coverage of important features in areas with scarce landmarks.
We tested our approach using a map generated using detected pole and corner features as inputs to an Extended Kalman filter (EKF) for simultaneous localisation and mapping \cite{slam}. The map was used to localise during subsequent drives, and the testing was carried out using the new data.
Our evaluation method consists of localising the vehicle within the resulting map and calculating the performance of the localisation algorithm by computing the magnitude of the maximum vehicle state covariance. 
In the same way, the map downsampling was assessed by the selective dropping of landmarks (based on our score) from the original map. 

In the next section, we introduce a review of the related work for map updating through the removal of features that no longer exist. 
In Section III, we describe the components of our approach and the methodology used to build the model. 
The experiments, evaluation and outcomes are presented in Section IV.

\section{Related work}

Changes in the perceived environment can be caused by environmental conditions (weather, trees foliage due to the seasons), time of day (shadows, illumination) and structural variations (new/demolished buildings, dynamic objects taken into account when the map was made). Some of these changes can be managed by just adding new information into the map, and relying on the data association algorithm to ignore outliers produced by these changes. 

Different works have addressed the problem of removing dynamic or non-existing features from feature maps in both vision-based and lidar-based systems. 
For visual localisation, to estimate the robot pose within an existing map it is necessary to retrieve the most reliable features (due to and bandwidth limitations) in the local map from the map storage to calculate the current pose \cite{Lynen2015GetOO}.
A diverse range of approaches have been used in order to reduce the size of a feature map. In \cite{Muhlfellner794800}, the authors presented an approach to choosing landmarks that are considered valuable for localisation based on statistical measures. In this work, each landmark is scored as a function of the times it has been observed. In \cite{7139575}, an algorithm that alternates between on-line place recognition and off-line map maintenance is presented and evaluated with the aim of producing a reliable fixed size map. The same authors in \cite{M_dymczyk_2016} propose a method for reducing the amount of map data that needs to be transferred by making use of predictors that include the distance travelled while observing a landmark, mean re-projection error and the classification of the descriptor appearance. Stability of visual features has also been explored, e.g. by assessing their distinctiveness and the robustness of image descriptors \cite{CARNEIRO20091143}, or uniqueness of the descriptors \cite{confVerdieYFL15}.

A different approach is presented in \cite{Meyer-Delius:2012:OGM:2900929.2901014} where the dynamic environment is represented by occupancy grids. Each cell has an associated state transition probability.
Another strategy for removing old objects is based on associating a persistence time measure to each feature. This approach assigns a set life time to each feature each time it is re-observed \cite{7487237}. If by the end of this time span the feature has not been matched with any observation, it is removed from the map \cite{lifetime_f}. 
A recent work described as frequency map enhancement \cite{7759671,7878680} develops a spatio-temporal representation to describe the persistence and periodicity of the individual cells/features allowing the future calculation of an occupancy/occurrence prediction probability.

In contrast to the visual based mapping approaches, our model is generic and can be used to map any kind of feature/landmark in 2D or 3D. This is because it is based on predictors that are not attached to specific feature properties, instead based only on the geometry as measured from the dataset. This methodology was validated with real data using a prior 2D map.

\section{Methodology}

\begin{figure}[h]
\vspace{3mm}
   \centering
   \includegraphics[scale=.23]{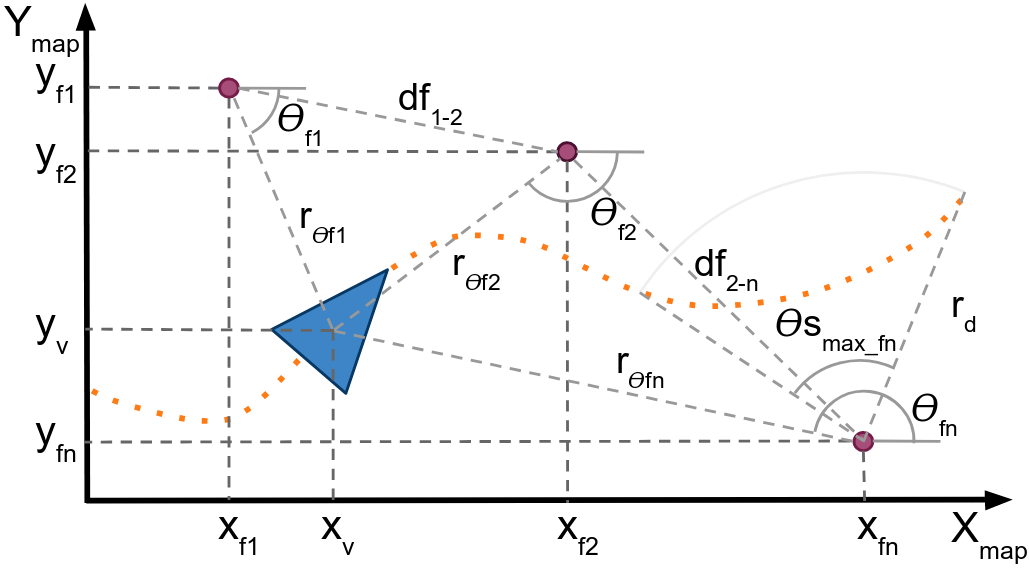}
    \caption{The vehicle $v$ navigates within the featured-map $[x_{map}, y_{map}]$ composed by n landmarks, located in coordinates $[x_{fn}, y_{fn}]$ }
    \label{fig:V_model}
\end{figure}

The measure of importance given to a landmark for localisation purposes is established based on a combination of different variables. The objective of this paper is to derive, examine, analyse and evaluate these importance variables. The variables are related to the observability and distribution of the landmarks, leading to a relevance score.

Different predictors are proposed, selected, adjusted and later combined to evaluate the capacity of each feature to be used for localisation in future datasets. We defined some notations as follows: 

\begin{itemize}

\item $[x_{fn}, y_{fn}]$ denotes the location of the landmark $n$ in the map global frame.

\item  $[x_{v}, y_{v}]$ denotes the location of the vehicle in the map global frame. 

\item $\theta_{fn}$ is the detection angle of the landmark $n$ in its own local frame when the vehicle is located in $[x_{v}, y_{v}]$.

\item  $r_{\theta fn}$ is the distance between the vehicle and the landmark when the last time it was detected at the angle $\theta_{fn}$.

\item   $\theta s_{max_fn}$ denotes the spanned angle of the landmark detection based on the vehicle's trajectory.

\item   $df_{n-m}$ denotes the distance between the landmarks $n$ and $m$ respectively.  

\end{itemize}

\subsection{Predictor variables}

\subsubsection{Number of detections} 
By considering variables associated with the feature's influence on the quality of localisation and its persistence during the vehicle's operation, it is clear that the number of times $n_{F_n}$ the landmark $n$ has been observed is directly related to the usefulness of the feature.
This predictor has been used extensively in map downsampling approaches \cite{7054429,Muhlfellner794800}. The value as a predictor is based on the assumption that the more matches of the landmark $n$, the greater the expectation that it will be detected again and consequently be used by the localisation algorithm. 
Any time a landmark is matched to the prior map, the corresponding feature counter $n_{F_n}$ is increased by one, regardless of the vehicle's position or speed.

\subsubsection{Maximum detected spanning angle} 
This variable corresponds to the maximum angle covered by all detections/matches achieved from the moving vehicle to each landmark. 
The fact the landmark can be seen from a comparably extended set of angles illustrates the relevance of that landmark to localize the vehicle in the surrounding area. To every landmark, a vector of discretized angles $\theta_{dfn}$ is assigned, in our case with $1^\circ$ resolution, at the moment the landmark is successfully matched, the angle to the vehicle $\theta_{fn}$ is calculated and rounded, and the corresponding vector position is marked indicating that a successful feature match was achieved from that angle. 
Thus, the maximum detected spanned angle $\theta d_{m_fn}$ of the landmark $n$ is calculated as the sum of all components within $\theta_{dfn}$.

\begin{equation} \label{eq1}
\theta d_{m_fn} = sum(\theta_{dfn})
\end{equation}

\subsubsection{Maximum length driven while observing the landmark} 
A predictor derived from $\theta d_{m_fn}$ is the maximum length driven while the landmark is being observed $Ml_{fn}$. The length of the path from where the landmark can be observed is taken into consideration. The $\theta d_{m_fn}$ also depends on the environment structure (open area or with the presence of several obstacles) and the vehicle's trajectory (rectilinear or curvilinear) within the map. Similarly to the calculation of $\theta d_{max_fn}$, a vector of ranges associated to each discrete angle $mR_{fn}$ is designated to each landmark. When the landmark is re-observed, the distance from the vehicle to the landmark $r_{\theta fn} $ is computed and compared with any previous value belonging to the same detection angle $\theta_{fn}$. The maximum value is then registered in the vector position corresponding to $\theta_{fn}$. The $Ml_{fn}$ is determined as the length of the arc described by the vehicle. This is calculated as the dot product between the vector $mR_{fn}$ and the vector of discrete unit angles $A_r$ for which all components are equal to $pi/180$ ($1^\circ$ angle resolution).

\begin{equation} \label{eq1}
Ml_{fn} =  mR_{fn}\cdot A_r
\end{equation}

\subsubsection{Maximum area of detection}
In a similar manner, the Maximum area of detection $Ma_{fn}$ was calculated. In this predictor we compute the total detection area of the landmark by assuming that it could be detected by any of the geographical points between it's position $(x_{fn}, y_{fn})$ and the vehicle's pose $(x_v,y_v)$. $Ma_{fn}$ is estimated as half the dot product between the pairwise power of the vector $mR_{fn}$ and $A_r$. 

\begin{equation} \label{eq1}
Ml_{fn} =  \frac{1}{2} mR_{fn}^2\cdot A_r
\end{equation}

\begin{figure}[t!]
\vspace{3mm}
\centering

\begin{subfigure}[]{0.32\columnwidth}
\centering
	\includegraphics[width=\columnwidth]{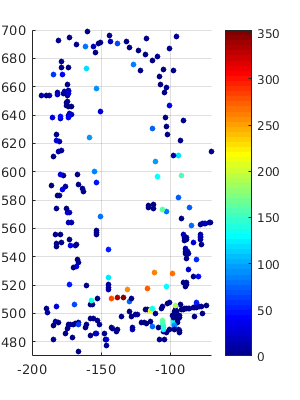}
    \caption{Detections.}
    \label{sub_a}
    \end{subfigure}
\begin{subfigure}[]{0.32\columnwidth}
\centering
	\includegraphics[width=\columnwidth]{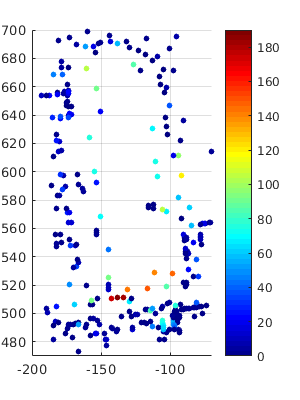}
    \caption{S angle.}
    \label{sub_b}
    \end{subfigure}
\begin{subfigure}[]{0.32\columnwidth}
\centering
	\includegraphics[width=\columnwidth]{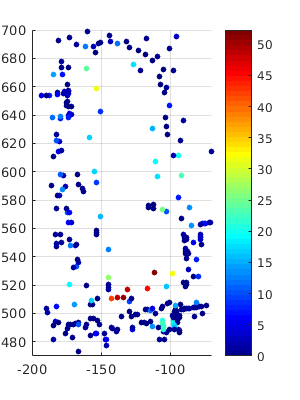}
    \caption{Track length.}
    \label{sub_c}
    \end{subfigure}
    
\begin{subfigure}[]{0.32\columnwidth}
\centering
	\includegraphics[width=\columnwidth]{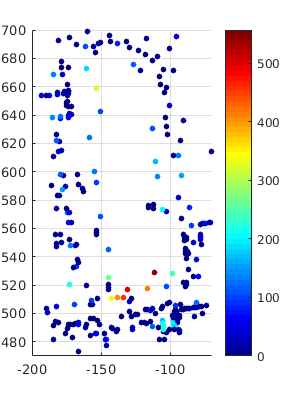}
    \caption{Area.}
    \label{sub_d}
    \end{subfigure}    
\begin{subfigure}[]{0.32\columnwidth}
\centering
	\includegraphics[width=\columnwidth]{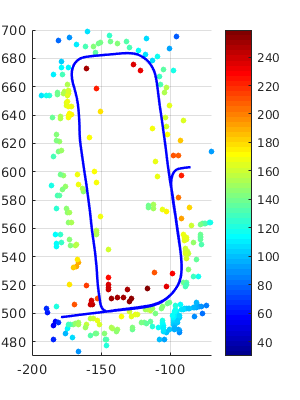}
    \caption{M.S. angle.}
    \label{sub_e}
    \end{subfigure}
\begin{subfigure}[]{0.32\columnwidth}
\centering
	\includegraphics[width=\columnwidth]{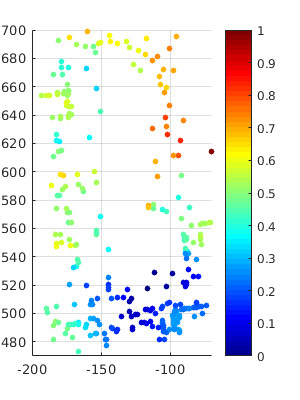}
    \caption{C. ratio.}
    \label{sub_f}
    \end{subfigure}

\caption{\small Predictor variables. \ref{sub_a}. Number of detections per landmark. \ref{sub_b}. Maximum detected spanned angle. \ref{sub_c}. Maximum length driven while observing the landmark. \ref{sub_d}. Maximum area of landmark detection. \ref{sub_e}. Maximum possible spanned angle. \ref{sub_f}. Concentration ratio.}
\label{fig:proceso}
\end{figure}

\subsubsection{Maximum possible spanned angle } 
Per every feature $n$ and by looking for the vehicle's vector poses within a radius of 30 meters on the feature frame, we calculated the maximum possible spanned angle $\theta s_{max_fn}$. This predictor depicts the landmark's potentiality for localisation. It is not susceptible to environmental changes that may occlude the landmark since it is assumed that the vehicle navigates in an open space, hence  $\theta s_{max_fn}$ depends only on the vehicle's trajectory.

\subsubsection{Concentration ratio} 
The spatial distribution of landmarks within the map should be considered in any feature map generation algorithm. 
Areas with few landmarks for localisation should keep as many of the features as possible, whereas areas with dense features can be reduced.
To measure the weight of each landmark within its surroundings, we introduce a predictor called concentration ratio $Cr_n$. 
The concentration ratio for a given feature indicates the density of surrounding features used for localisation.
To calculate the concentration ratio $Cr_n$ for every landmark $n$ in the map, all distances $df_{n-i}$ to the landmarks $i$ located within a 30 metre radius of $n$ are calculated. The value of $Cr_n$ is then estimated as the division between the furthest landmark $max(df_{n-i})$ from $n$ and the sum of the distances $df_{n-i}$. 

\begin{equation} \label{eq1}
Cr_n =  \frac{max(df_{n})}{\sum{df_{n-i}}}
\end{equation}

A low value of $Cr_n$ indicates a higher density of features for localisation compared with a concentration ratio approaching 1 which indicates the landmark density is sparse.

\subsection{Scoring function} 

The scoring function aims to estimate the relevance to localisation for each landmark across a number of datasets as a function of each of the predictors.
We propose to use a regression algorithm to adjust the coefficients of the identified predictors calculated using a training dataset classified by its empirical probability.
The empirical probability function can be defined as the vector of frequencies of each landmark detection when considering the dataset \cite{Empirical_prob}. 
This vector is normalized using the percentage of the datasets in which the landmark has been re-observed.
Each dataset was registered in the same area used to create the initial feature map. 
As the vehicle was driven approximately along the same trajectory each time, persistent landmarks from the map should be recognized.
The localisation algorithm was modified to retrieve the details of the matched landmarks each time a detection was performed.

The predictors need to be standardised before they can be incorporated into the scoring function. This is required as the scale of the predictors varies and for the regression model we are proposing, normalisation of the scale is needed for each parameter to contribute equally.
After this pre-processing step, we perform use a ridge and lasso model to minimise the regression coefficients. 
This is based on a sum of the squared coefficients and the sum of the absolute coefficients.
The predictors with lesser significance to the result have the corresponding coefficients set to near zero.
It is critical for the analysis of the predictors that they can be selected, removed or transformed to formulate a reliable model.

After the predictor coefficients were selected, we use a cross-validated elastic net regularized regression method \cite{Zou05regularizationand} to fit our model. The method is switched between Lasso (L1) and Ridge (L2) regularization methods to overcome the limitations of each approach based on an alpha parameter. 
Elastic net is suitable for this case where some predictors are strongly correlated.

\section{Results}

\begin{figure}[h!]
   \centering
   \includegraphics[scale=.32]{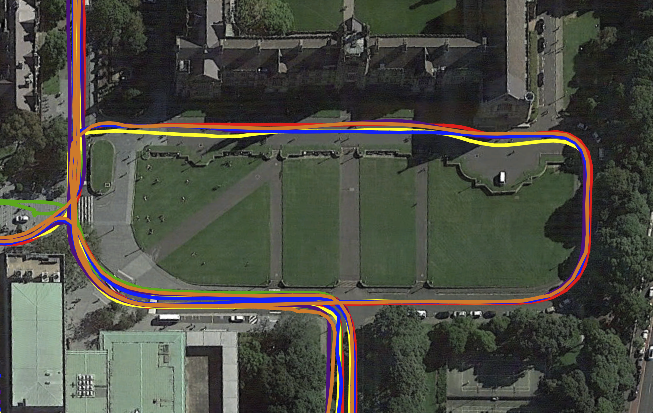}
    \caption{A 500m track length dataset was captured once per week over a six month period at different times of day.}
    \label{fig:ssmap}
\end{figure}

We tested our approach using a prior feature map generated from a single dataset Fig. \ref{fig:ssmap}. An electric vehicle (EV) equipped with a 16 beam Velodyne laser sensor was driven in a clockwise direction in front of the main quadrangle of the University of Sydney to collect the dataset used to build the initial feature-map.
No special conditions were enforced for the area of the collected dataset, meaning vehicles, pedestrians and cyclists were also observed by the laser. The length of the track was around 500 metres.

The features extracted from the point cloud corresponds to two feature categories, poles and corners, which are presumed to belong mainly to static objects.
However, corners of parked and moving vehicles (especially trucks) and pedestrians were sometimes detected as poles and included as observations. An extended EKF-SLAM algorithm was used to build the prior feature map shown in Fig. \ref{fig:mapa}.

Over the course of six months, data collection was performed around the University of Sydney campus on weekly basis on different days of the week and hours of the day. During this period different construction works were carried out in the mapped area, causing the disappearance of objects in the zone. This information obtained over this half year time period was used to build and test our model.

The localisation algorithm using the prior map estimated the global pose of the vehicle based on an iterative closest point (ICP) data association algorithm, which identified the pose corresponding to the current observations. An off-line algorithm was implemented to compute each of the predictors based on the localisation algorithm output (vehicle pose and observed features).

For predictor selection we used ridge and lasso regularisation models, and checked which coefficients were set to zero. For the lasso case, the coefficient belonging to the number of detections of each landmark was set to zero due to its high correlation with other predictors, having a maximum correlation coefficient $R^2$ of 0.96 with the detected spanned angle.
For the ridge model, the coefficients set to zero corresponded to Maximum possible spanned angle and the concentration ratio. Since we want our model to include the concentration ratio so that areas with a lower density of landmarks will be maintained, we used a non-linear transformation formed by the multiplication of the coefficient ratio and the number of detections. 

Table \ref{tab:predictors} shows the $R^2$ between the predictors and the empirical probability used to label the data.

\begin{table}[h]
\centering
\begin{tabular}{|l|l|l|l|}
\hline
\multicolumn{1}{|c|}{\textbf{Predictor}} & \multicolumn{1}{c|}{\textbf{$R^2$}} & \multicolumn{1}{c|}{\textbf{Predictor}} & \multicolumn{1}{c|}{\textbf{$R^2$}} \\ \hline
Number of views                          & 0.2352                              & Maximum Angle                           & 0.0599                              \\ \hline
Spanned Angle                            & 0.2947                              & CR                                      & 0.0184                              \\ \hline
Track Length                             & 0.2805                              & CR*N Views                       & 0.2523                              \\ \hline
Total Area                               & 0.2562                              &                                         &                                     \\ \hline
\end{tabular}
\caption{Predictors' coefficients of correlation with the labelled output.}\label{tab:predictors}
\end{table}

Having selected and transformed the predictors, we used a cross-validated elastic net regularized regression method to tune the $\alpha$ parameter and select the coefficients corresponding to $\lambda$ with minimum cross-validation error plus one standard deviation. The achieved coefficient of correlation $R^2=0.76$, which implies that three quarters of the variance is related to the predictors.

\begin{figure}[h]
   \centering
   \includegraphics[scale=.47]{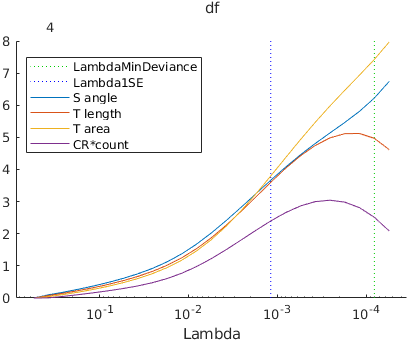}
    \caption{Trace plot of coefficients fit by elastic net. The blue dotted line denotes the $\lambda$ with minimum cross-validation error plus one standard deviation.}
    \label{fig:V_model_coeff}
\end{figure}

To select the most reliable landmarks that will comprise our final map, we fit a non-parametric kernel-smoothing distribution to the predictions histogram Fig.\ref{fig:V_model}. 
This allows us to obtain the local minimum point between the extreme peaks and select those features above the difference between the identified local minima and 0.5 times its standard deviation. This last component was included in order to incorporate features which could be occluded during data collection, or could not be detected by the perception system during a particular dataset. 

\begin{figure}[h]
   \centering
   \includegraphics[scale=.5]{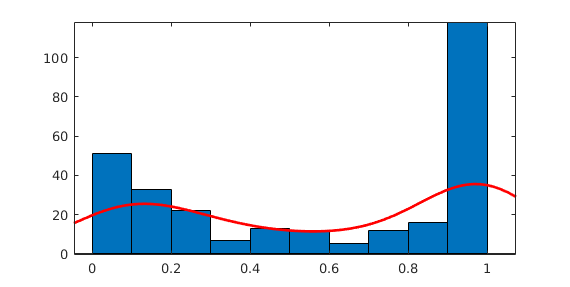}
    \caption{Predictions histogram and distribution.}
    \label{fig:V_model}
\end{figure}

\begin{figure}[h]
   \centering
   \includegraphics[scale=.28]{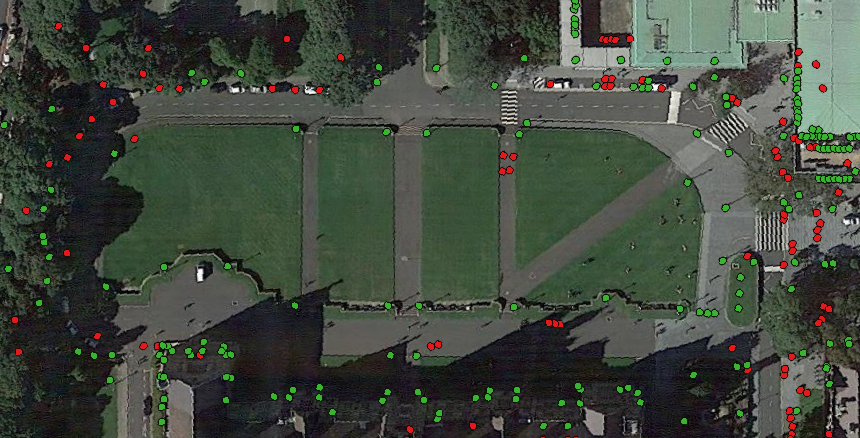}
    \caption{Map built with identified static and semi-static landmarks in green, discarded landmarks in red.}
    \label{fig:final_map}
\end{figure}

A map with the previous selected landmarks, Fig.\ref{fig:final_map}, has $35\%$ less features than the initial map. Through a visual inspection of the zone and the map itself, we corroborated the correspondence to the ground truth.
Our approach was able to identify and remove features in the middle of the road, pedestrian paths, parking areas, and other features difficult to identify by the sensors. Also, landmarks that were very high from the ground and inside of buildings were likely to be pruned.

\begin{figure}[h]
\vspace{3mm}
   \centering
   \includegraphics[scale=.65]{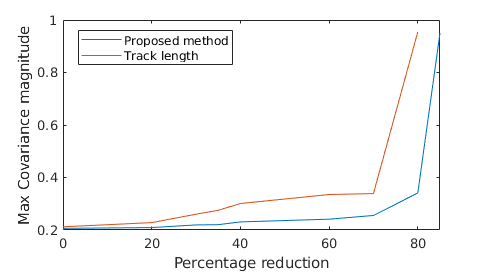}
    \caption{Localisation covariance versus feature drop rate.}
    \label{fig:cov_res}
\end{figure}

Since the regression predictions scores the landmarks for localisation, we tested our downsampling capability and compare it with the approach in \cite{7139575} which only takes into account the travelled distance while observing the landmark. 
The dataset used for the experimental results come from the later weeks in the half year of weekly datasets, and therefore was not included in the initial model fitting.
Fig.\ref{fig:cov_res} show the outcome of this comparison.
We started by dropping the least valuable $20\%$ of the features. 
The maximum covariance magnitude remains the roughly same for both methods. After $20\%$ drop rate this value for our proposed method is always below the "distance travelled while observing" metric, and is almost constant until a drop rate of $70\%$. 
It is important to note that the evaluation of the final map in Fig.\ref{fig:final_map} corresponds to $35\%$ drop rate, which is where the maximum covariance magnitude shows a negligible increment in contrast to the entire map.

Both methods will lead to failures of the localisation algorithm once the feature map is very sparse. 
This failure occurs when the drop percentage is $75\%$ for the method based on the track length and $85\%$ for the proposed method. This demonstrates that our method is better at predicting the importance of features for localisation and can correctly prioritise the removal of less important features.

\section{Conclusion and future work}
In this paper we presented an approach to evaluate landmarks based on the likelihood of being persistently re-observed over future visits to the mapped area. Our approach allows us not only to discard unstable landmarks and have a map more suited to perform localisation, but also score each landmark so that the map can be reduced while maintaining reliable localisation. 

The scoring method is based on an elastic net regression model which incorporates a range of predictor variables that are related to the capability of each landmark to improve the localisation of the vehicle.
The feature concentration ratio was included as a novel predictor to allow the system to measure the degree of feature sparseness related to localisation in the neighbouring area and consequently give it a larger weight in the scoring function.
Even though maximum possible spanned angle predictor was not used in this model, we believe it could be a valuable predictor to include in cases where the vehicle/robot can travel the mapped area in two different directions. This is because the feature detector does not detect features uniformly within the sensor's coverage area, and that the environment was crowded with pedestrians and vehicles. In addition, there were various sources of occlusions that would commonly occur. This can result in the vehicle travelling within the map but not estimating the same track that was travelled when the original map was built.
For this paper, the data collection was done in a single direction and this predictor was discarded. In future work we intend to include this metric in order to assess the contribution when the vehicle is operated in the cases mentioned.

We demonstrate the downsampling ability of the model by comparing its performance with a state of the art model. Our model achieves better results for localisation purposes at all downsampling ratios.

\addtolength{\textheight}{-12cm}   




\section*{ACKNOWLEDGMENT}

This work has been funded by the ACFR, the University of Sydney through the Dean of Engineering and Information Technologies PhD Scholarship (South America) and the Australian Research Council Discovery Grant DP160104081 and University of Michigan / Ford Motors Company Contract ``Next generation Vehicles".



\bibliography{main}
\bibliographystyle{IEEEtran}


\end{document}